\PassOptionsToPackage{table,xcdraw}{xcolor}
\documentclass[sigconf]{acmart}
\renewcommand\footnotetextcopyrightpermission[1]{} 
\AtBeginDocument{%
  }

\usepackage{multirow}
\usepackage{subfigure}
\usepackage{colortbl}
\usepackage{graphicx}
\usepackage{booktabs}
\usepackage{adjustbox}
\setcopyright{none}

\begin{document}
\pagestyle{plain}

\title{A Step towards Interpretable Multimodal AI Models with MultiFIX}

\author{Mafalda Malafaia}
\affiliation{%
  \institution{Centrum Wiskunde \& Informatica}
  \city{Amsterdam}
  \country{The Netherlands}
}
\email{Mafalda.Malafaia@cwi.nl}
\orcid{0000-0002-8081-0454}

\author{Thalea Schlender}
\affiliation{%
  \institution{Leiden University Medical Center}
  \city{Leiden}
  \country{The Netherlands}
}
\email{T.Schlender@lumc.nl}
\orcid{0000-0001-7385-112X}

\author{Tanja Alderliesten}
\affiliation{%
  \institution{Leiden University Medical Center}
  \city{Leiden}
  \country{The Netherlands}
}
\email{T.Alderliesten@lumc.nl}
\orcid{0000-0003-4261-7511}

\author{Peter A. N. Bosman}
\affiliation{%
  \institution{Centrum Wiskunde \& Informatica}
  \city{Amsterdam}
  \country{The Netherlands}
}
\affiliation{
  \institution{Delft University of Technology}
  \city{Delft}
  \country{The Netherlands}
}
\email{Peter.Bosman@cwi.nl}
\orcid{0000-0002-4186-6666}

\renewcommand{\shortauthors}{Malafaia et al.}

\begin{abstract}
Real-world problems are often dependent on multiple data modalities, making multimodal fusion essential for leveraging diverse information sources. In high-stakes domains, such as in healthcare, understanding how each modality contributes to the prediction is critical to ensure trustworthy and interpretable AI models. We present MultiFIX, an interpretability-driven multimodal data fusion pipeline that explicitly engineers distinct features from different modalities and combines them to make the final prediction. Initially, only deep learning components are used to train a model from data. The black-box (deep learning) components are subsequently either explained using post-hoc methods such as Grad-CAM for images or fully replaced by interpretable blocks, namely symbolic expressions for tabular data, resulting in an explainable model.  We study the use of MultiFIX using several training strategies for feature extraction and predictive modeling. Besides highlighting strengths and weaknesses of MultiFIX, experiments on a variety of synthetic datasets with varying degrees of interaction between modalities demonstrate that MultiFIX can generate multimodal models that can be used to accurately explain both the extracted features and their integration without compromising predictive performance.
\end{abstract}




\maketitle

\vspace{-4mm}
\section{Introduction}
\label{sec:intro}


Data availability is significantly expanding across numerous domains, not only in volume but also in diversity, translating to a heterogeneous data landscape~\cite{arora2023introduction}. Contrary to the oftentimes unimodal nature of Artificial Intelligence~(AI) approaches, domain experts rely on multiple data modalities in their decision-making processes. For instance, in the healthcare domain, medical experts typically consider medical imaging exams, demographics, blood analysis, and further clinical information to make an informed decision. It is known that Multimodal Machine Learning~(ML) can outperform single-modality approaches~\cite{rahate2022multimodal}, offering increased robustness and the ability to leverage complementary information~\cite{kline2022multimodal}.

Despite the state-of-the-art performance of Deep Neural Networks~(DNNs) in various unimodal and multimodal tasks~\cite{zhao2024deep}, their opaque nature can present challenges in high-stakes domains, where interpretability and trust are paramount. Thus, to be employed in real-world situations, AI frameworks must be human-verifiable, and in some cases interpretable~\cite{rudin2022interpretable}, considering not only ethical but also legal and privacy aspects. 

Interpretable multimodal approaches increase transparency, can lead to knowledge discovery, and enable verifiability. Moreover, these promote vital and constant interaction between AI and domain experts, interpreting how models work, and providing input on if and how models should be adjusted~\cite{joshi2021review}.
Especially for predictive models in several high-stakes fields, this is of key importance.

In this work, we demonstrate MultiFIX: a Multimodal Feature engIneering approach to eXplainable AI. MultiFIX is a framework that is aimed at interpretability through the discovery of key features for different modalities, the combination of which is used to make predictions. By providing explanations of the features and the final predictions, MultiFIX provides a unique, novel approach to explainable multimodal AI. To develop MultiFIX models, the powerful learning potential of Deep Learning~(DL) and the interpretability of symbolic expressions generated with Genetic Programming~(GP) are leveraged. The latter is used to create models that can readily be analyzed and interpreted as a surrogate for the deep learning model. Additionally, other modality-specific post-hoc explanation techniques can be used to study model components in MultiFIX - for instance, the use of Grad-CAM for image processing neural networks.


Additionally to the MultiFIX pipeline, a key contribution of this paper is the examination of different training strategies. Preliminary work on MultiFIX~\cite{malafaia2024multifix}  illustrates initial favorable outcomes in multimodal integration, primarily using end-to-end training.
While end-to-end training has the potential to effectively leverage joint optimization of all components in MultiFIX, the complexity of the joint learning task may present certain limitations. We therefore also consider sequential and hybrid training, in combination with different pre-training procedures, providing flexibility in the most suitable training strategy according to the nature of the problem and the preferred architectural blocks in practice. This paper provides the first comprehensive analysis of the MultiFIX pipeline and various training methods, making it the most complete and detailed study on the subject.
To demonstrate the versatility and potential of MultiFIX, in this paper we perform an exploratory study on synthetic problems with various degrees of dependence between modalities that are representative of real-world scenarios.

The remainder of this paper is organized as follows: Section \ref{sec:related_work} reviews related work; Section \ref{sec:methods} details the MultiFIX methodology and experimental design; Section \ref{sec:results} presents the staged problems and outcomes, namely the resulting interpretable models; Section \ref{sec:discussion} reflects a discussion on the exploratory study; and Section \ref{sec:conclusion} concludes the present work with a discussion of future directions.

\vspace{-2mm}
\section{Related Work}
\label{sec:related_work}

We specifically focus on techniques that combine image and tabular data - two modalities that are commonly incorporated in multimodal AI pipelines~\cite{sleeman2022multimodal}.

Multimodal learning methods are commonly categorized by fusion strategies: early fusion, which concatenates input data before feature extraction; intermediate fusion, which combines modality-specific features after extraction; and late fusion, which aggregates predictions from unimodal models~\cite{huang2020fusion,stahlschmidt2022multimodal}. Late fusion is the dominant strategy due to its simplicity and effectiveness in handling heterogeneous data~\cite{sleeman2022multimodal}. However, recent advancements in intermediate fusion methods provide enhanced ways of capturing interactions between modalities, especially with the development of DL techniques for feature extraction~\cite{guarrasi2024systematic}. 

Despite these advancements, most multimodal systems still lack interpretability, a crucial aspect for building trust in AI. Recent literature highlights the importance of interpretability in multimodal models, particularly in high-stakes sectors where transparency and clarity on how the models work are critical to using them~\cite{schouten2024navigating}. Current state-of-the-art literature on explainable multimodal approaches relies substantially on modality-specific post-hoc methods. 

Post-hoc explainability methods for image data have been primarily used in unimodal approaches, where the exclusive contribution of the image is mapped to the prediction. Grad-CAM is arguably the most-used method among gradient-based methods that rely on the backpropagation process of Convolutional Neural Networks~(CNNs) to generate attention maps in the images~\cite{allgaier2023grad}.

 The most common explainability method for tabular data analysis is the post-hoc use of SHAP values to analyze feature importance~\cite{allgaier2023grad}. Another approach is GP, which can be used to evolve higher-level features in the form of symbolic expressions~\cite{virgolin2020explaining,zhou2023evolutionary}, or to directly evolve prediction models that are fully white-box~\cite{bacardit2022intersection}. In comparison to post-hoc methods like SHAP, GP directly generates interpretable models without additional approximations, reducing the risk of misleading explanations. In Evans et al.~\cite{evans2019s}, GP is used to generate compact symbolic expressions that are used as a post-hoc method to approximate ML estimators without compromising the predictive performance significantly.

Utilizing single modality, post-hoc, explainability methods, Chen et al.~\cite{chen2022pan} propose an explainable multimodal pipeline using an intermediate fusion approach to provide prognostic predictions across 14 types of cancer. Specifically, they use attention mechanisms for histology images and SHAP values for genomic data to find correlations between input features and the target prediction.

Post hoc explanations are frequently chosen to explain black-box models, but may provide misleading and unreliable explanations~\cite{rudin2019stop}. Hence, Swamy et al.~\cite{swamy2024multimodn} describe an inherently interpretable multimodal approach that indicates the cumulative contributions of each modality to the prediction. Additionally, the modularity of the pipeline allows multi-task predictions while handling potentially missing modalities, due to its sequential training strategy. However, the proposed pipeline does not include interpretability on a feature extraction level, i.e., the feature contributions within each modality remain a black box, which arguably limits explainability, including knowledge discovery.

In contrast to existing literature, MultiFIX introduces innovative feature engineering interpretability with explicit contributions of each modality to the final prediction. Specifically, we address two major challenges: to incorporate \emph{inherently} interpretable fusion techniques within an intermediate fusion pipeline by using GP to generate symbolic expressions; and to use sparse embedded feature engineering to extract a narrow bottleneck of modality-specific features that capture patterns that are potentially relevant to the prediction.

\vspace{-3mm}
\section{Methodology}
\label{sec:methods}

We describe the MultiFIX pipeline, including methods used and the experimental setup. Additionally, we provide a general description of the input data used.

\subsection{MultiFIX: Multimodal Feature engIneering for eXplainable AI}

In MultiFIX, DL architectures are trained to generate black-box models that are designed to learn representative, potentially complex features from each modality. Whenever possible, these models are replaced by GP-generated symbolic expressions that are interpretable by design, and otherwise are explained using Grad-CAM to generate visual explanations. While we use Grad-CAM in this work, other image explainability methods could be used.

\begin{figure}[b]
    \centering 
    \includegraphics[width=0.49\textwidth]{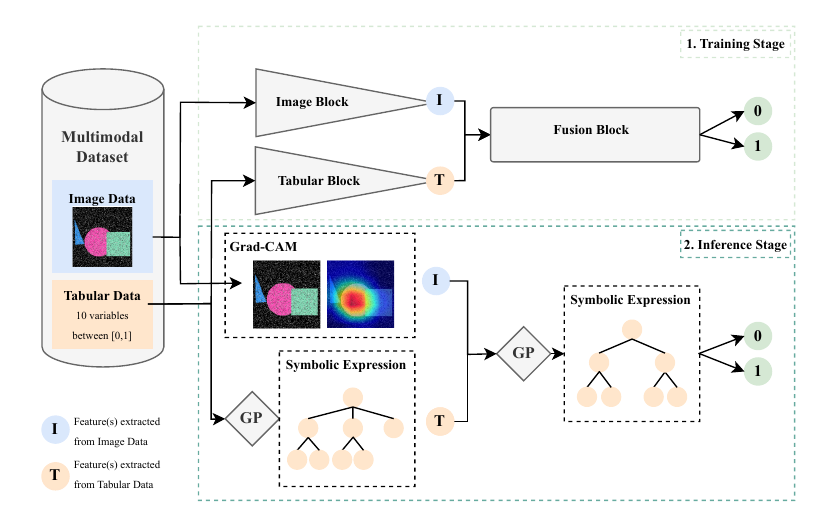}
    \caption{Overview of MultiFIX. Data passes into the feature engineering blocks. Feature vectors I and T are concatenated and passed to the fusion block to make the final prediction in the Training Stage~(top). In the Inference Stage, image features are explained through Grad-CAM, and symbolic expressions are obtained for both the tabular features and the target prediction with GP-GOMEA, replacing their NN counterparts.}
    \label{fig:pipeline}
\end{figure}

We demonstrate our pipeline using two data modalities: image and tabular data. However, the modularity of the pipeline supports straightforward integration of other data modalities by adding modality-specific feature engineering blocks. An overview of MultiFIX is provided in Figure~\ref{fig:pipeline}.

\subsection{Feature Engineering Blocks}
The structure of each feature engineering block is very similar, regardless of the modality to be analyzed, comprising a DL architecture to process the input data and extract meaningful features. We introduce the concept of sparse feature engineering, where the feature engineering bottleneck is deliberately restricted to a small number of features (we use at most three in this paper), in order to increase the chance of obtaining (easily) interpretable models.

For image data, we use a Convolutional Neural Network~(CNN). In this paper, we use a pre-trained Resnet~\cite{resnet18}, but the architecture may be adjusted to suit specific needs or be optimized using neural architecture search. Besides this, an autoencoder (AE) can be trained (with the same architecture) and used as a pre-trained image processing block in some training strategies.

For tabular data, we use a Multi-Layer Perceptron~(MLP)  with three hidden layers, all of which are 128 nodes wide. Dropout and batch normalization are used to prevent overfitting and promote fast convergence.

\vspace{-2mm}
\subsection{Fusion Strategy}

MultiFIX employs an intermediate fusion strategy to combine the engineered features from each modality to make the final prediction. However, the unique enforcement of a bottleneck with up to three features per modality makes the fusion analysis simpler and, thus, interpretability-focused.

The architecture used for the fusion block in this paper consists of the same MLP architecture used for tabular processing. Here also, alternative architectures could be used, or architecture search could be applied.

\subsection{Training Strategies}
\label{subsec:training_strategies}
We study different training strategies with MultiFIX, to explore the potential benefits of different strategies specific to different types of problems and synergies between modalities. Additionally, single-modality approaches were used as a baseline comparison, as well as pre-trained blocks for the respective modalities.

Six different training strategies were considered, each varying either in pre-training weights for feature engineering blocks, sequential or parallel training, or partial temporary freezing of the architectural blocks:
\begin{enumerate}
    \item \emph{End-to-end training~(End):} train the entire architecture simultaneously.
    \item \emph{Sequential training with AE weights~(Seq AE):} use encoder weights from the trained AE in the image feature engineering block; freeze the latter while training the tabular feature engineering and the fusion blocks.
    \item \emph{Sequential training with AE weights and De-freezing~(Seq AE Temp Freeze):} use encoder weights from the trained AE in the image feature engineering block; freeze image block for 15 epochs while training remaining blocks and then train the whole architecture simultaneously.
    \item \emph{Sequential training with single modality weights~(Seq Single):} use weights from each single modality model for respective feature engineering blocks; train fusion block sequentially, while freezing the remaining blocks.
    \item \emph{Hybrid training with AE weights~(Hyb AE):} use encoder weights from the trained AE in the image feature engineering block; train the whole architecture simultaneously using the pre-trained block.
    \item \emph{Hybrid training with single modality weights~(Hyb Single):} use weights for each single modality model for respective feature engineering blocks; train whole architecture simultaneously using the pre-trained blocks.
\end{enumerate}

\subsection{Interpretability Techniques}

In the inference stage, the DL model blocks are either explained using post-hoc explainable methods or replaced by a symbolic expression obtained using inherently interpretable methods. For images, Grad-CAM~\cite{grad_cam19} leverages the gradient information from convolutional layers to generate visualizations that portray which parts of an image contribute (more) to a specific prediction. GP-GOMEA~\cite{gp_gomea21} is a model-based evolutionary algorithm for GP known for its effectiveness in evolving small and potentially interpretable symbolic expressions~\cite{srbench}. We use a recent adaptation of GP-GOMEA that allows the use of both numeric and Boolean operators, as well as if-then-else statements~\cite{schlender2024improving}. 

The explainability method used for feature engineering blocks depends on the nature of the data: Grad-CAM is used for images; GP-GOMEA is used for tabular data. GP-GOMEA is also used to replace the fusion block. Grad-CAM is applied to the activations from the last residual convolutional block of the ResNet, as suggested in guidelines~\cite{jacobgilpytorchcam}. For GP-GOMEA, we followed the settings described in Table 1 for all experiments.

\begin{table}[t]
\centering
\scalebox{0.8}{
\small
\begin{tabular}{l|l}
\textbf{Population size} & initially 64 (using IMS~\cite{gp_gomea21}) \\ \hline
\textbf{Number of generations } & 512 \\ \hline
                   & numeric $[+,-,*,/,.^2,.^3,]$\\
\textbf{Operators}  & Boolean $[==,\neq,>,<,AND,OR]$\\ 
                   & if-then-else \\ \hline
\textbf{Maximum tree depth} & {[}2, 3{]}
\end{tabular}
}
\caption{GP-GOMEA settings.}
\label{tab:gpg_settings}
\end{table}

\subsection{Experimental Setup}

The experiments were designed to evaluate the performance of the MultiFIX pipeline and the interpretability of the resulting models for different multimodal datasets and training strategies. Single-modality performance is used as a baseline to study whether improvements are obtained when data modalities are combined.

A standard experimental setup is used for all experiments, following the configurations described in Table~\ref{tab:opt_grid}. We use 5-fold cross-validation to assess the generalization capability of the models and a stratified 80/20 data split between train and validation sets. The Adam optimizer~\cite{kingma2015adam} is used with the Cross-Entropy~(CE) or Binary Cross-Entropy~(BCE) loss, depending on the nature of the target label. Early stopping is used with a patience of 5 epochs, with a maximum training period of 75 epochs. The batch size is 32. Hyper-Parameter Optimization~(HPO) is performed using a grid-search strategy to choose the optimal Learning Rate, Weight Decay, and number of extracted features~(up to three) for the image and tabular inputs (Image Bottleneck and Tabular Bottleneck, respectively). The optimal configuration can be unimodal if a bottleneck of zero features for either one of the modalities is chosen. This configuration will, however, differ from the unimodal baseline, since it also includes the fusion block, in this case combining features from a single modality.

For each of the studied problems, all training strategies mentioned in Subsection~\ref{subsec:training_strategies} are used, in addition to single modality approaches, resulting in eight trained models, all following the same protocol. The autoencoder, needed for three out of the six training strategies, is trained following the configuration of Table~\ref{tab:opt_grid}. The trained models are evaluated using Balanced Accuracy~(BAcc) for performance purposes, with loss and AUC-ROC metrics also being evaluated and available in the Supplementary Material. The final interpretable models are evaluated considering two aspects: predictive performance and interpretability. For the former, we compare BAcc values of the DL model with the interpretable model. Additionally, we compare the performance of the studied training strategies using statistical tests on cross-validation results. For each pair of strategies, we performed a paired t-test to determine if the performance differences were statistically significant. Following common statistical practice, we set the significance level at $\alpha=0.05$ and applied Bonferroni correction to control for inflated Type I error due to multiple pairwise comparisons. Interpretability is evaluated through a manual analysis.

\begin{table}[h]
\centering
\small
\scalebox{0.75}{
\begin{tabular}{c|ll}
AE settings
\multirow{4}{*}{\shortstack{\textbf{AE}\\\textbf{Settings}}} & \textbf{Optimiser} & Adam \\ \cline{2-3} 
& \textbf{Loss Function} & MSE \\ \cline{2-3} 
& \textbf{Learning Rate} & $0.0001$ \\ \cline{2-3} 
& \textbf{No. of Epochs} & $100$ \\ \hline
MultiFIX Settings
\multirow{4}{*}{\shortstack{\textbf{MultiFIX}\\\textbf{Settings}}} & \textbf{Optimiser} & Adam\\ \cline{2-3} 
& \textbf{Loss Function} & CE or BCE \\ \cline{2-3} 
& \textbf{No. of Epochs} & 75 \\ \cline{2-3}
& \textbf{Early Stopping Patience} & 5 epochs \\ \hline
Grid-Search
\multirow{4}{*}{\shortstack{\textbf{HPO}\\\textbf{Grid}}} & 
\textbf{Learning Rate} & {[}1e-3, 1e-4, 1e-5{]} \\ \cline{2-3} 
& \textbf{Weight Decay} & {[}1e-3, 1e-4, 0{]} \\ \cline{2-3}
& \textbf{Image Bottleneck} & {[}0, 1, 2, 3{]} \\ \cline{2-3}
& \textbf{Tabular Bottleneck} & {[}0, 1, 2, 3{]}
\end{tabular}}
\caption{Settings used in the MultiFIX pipeline. The best parameters are chosen according to the loss (lowest average $\pm$ standard deviation over the 5 folds).}
\label{tab:opt_grid}
\end{table}


\subsection{Dataset}

\begin{figure*}[]
    \centering
    \includegraphics[width=0.6\textwidth]{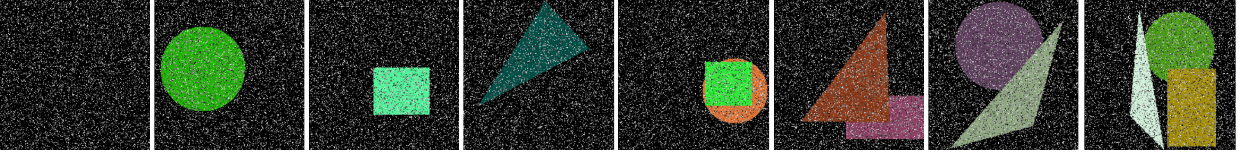}
    \caption{Representative samples for the image modality.}
    \label{fig:img_samples}
\end{figure*}

We created a synthetic dataset with images and tabular data. For the imaging modality, 1,000 samples were automatically generated, each with a size of $200\times200$ pixels. Each image can contain the following shapes: a circle, a rectangle and/or a triangle. Each shape can either be present or absent in the image, with the possibility of having none or up to three shapes. None of the shapes can appear twice in the same sample, and all are generated randomly with a $50\%$ chance, in different sizes and colors. Random noise was introduced to all images by mutating the color of 10,000 random pixels. An illustration of possible samples is presented in Figure~\ref{fig:img_samples}. Tabular data consists of 1,000 samples with ten numerical features, uniformly sampled between 0 and 1, that are then used in each problem to synthetically engineer tabular features that combine two or more of the numerical input features.

\section{Experiments and Results}
\label{sec:results}

In this section, we present the results of training the proposed MultiFIX pipeline using various training strategies for five synthetic problems that feature different dependencies between modalities.

Each subsection pertains to one of the proposed problems, including: the problem description; DL performance results for single modality approaches and six multimodal approaches with different training strategies; and the explainable model for the best performing strategy.

\subsection{AND Problem}

\subsubsection{Problem Description}

This problem comprises the binary operation AND between the presence of a circle~($circle$) in the image data, and the tabular feature $x_{1} > x_{2}$. The target label is thus the output of $AND(circle,x_{1} > x_{2})$. This problem shows a moderate level of dependence between the two modalities to predict the target, since the target label $1$ is only possible if both engineered features are present, but the target label $0$ correlates with any feature being $0$.

\subsubsection{DL Performance Results}

Figure~\ref{tab:and_bacc} presents the performance results for the AND Problem using a single modality and MultiFIX with different training strategies. Statistical testing highlights that all multimodal approaches significantly outperform single modality approaches. The multimodal approaches perform generally similarly from a statistical significance perspective, excluding the comparison between end-to-end and sequential AE approaches, in which the former is significantly better than the latter.

\begin{table*}[]
\centering
\begin{adjustbox}{max width=0.8\textwidth}
\begin{tabular}{lcccccccc}
\toprule
 & \textbf{Image Only} & \textbf{Tabular Only} & \textbf{End-to-End} & \textbf{Hybrid AE} & \textbf{Hybrid Single} & \textbf{Sequential AE} & \textbf{Sequential AE Defreeze} & \textbf{Sequential Single} \\
\midrule
\textbf{Image Only} & \cellcolor[HTML]{A9D0F5}0.607 ± 0.041 & not significant & \cellcolor[HTML]{F6B7B7} worse & \cellcolor[HTML]{F6B7B7} worse & \cellcolor[HTML]{F6B7B7} worse & \cellcolor[HTML]{F6B7B7} worse & \cellcolor[HTML]{F6B7B7} worse & \cellcolor[HTML]{F6B7B7} worse \\
\textbf{Tabular Only} & not significant & \cellcolor[HTML]{A9D0F5}0.693 ± 0.034 & \cellcolor[HTML]{F6B7B7} worse & \cellcolor[HTML]{F6B7B7} worse & \cellcolor[HTML]{F6B7B7} worse & \cellcolor[HTML]{F6B7B7} worse & \cellcolor[HTML]{F6B7B7} worse & \cellcolor[HTML]{F6B7B7} worse \\
\textbf{End-to-End} & \cellcolor[HTML]{C3F7C3} better & \cellcolor[HTML]{C3F7C3} better & \cellcolor[HTML]{A9D0F5}0.939 ± 0.029 & not significant & not significant & \cellcolor[HTML]{C3F7C3} better & not significant & not significant \\
\textbf{Hybrid AE} & \cellcolor[HTML]{C3F7C3} better & \cellcolor[HTML]{C3F7C3} better & not significant & \cellcolor[HTML]{A9D0F5}0.925 ± 0.022 & not significant & not significant & not significant & not significant \\
\textbf{Hybrid Single} & \cellcolor[HTML]{C3F7C3} better & \cellcolor[HTML]{C3F7C3} better & not significant & not significant & \cellcolor[HTML]{A9D0F5}0.923 ± 0.014 & not significant & not significant & not significant \\
\textbf{Sequential AE} & \cellcolor[HTML]{C3F7C3} better & \cellcolor[HTML]{C3F7C3} better & \cellcolor[HTML]{F6B7B7} worse & not significant & not significant & \cellcolor[HTML]{A9D0F5}0.881 ± 0.032 & not significant & not significant \\
\textbf{Sequential AE Defreeze} & \cellcolor[HTML]{C3F7C3} better & \cellcolor[HTML]{C3F7C3} better & not significant & not significant & not significant & not significant & \cellcolor[HTML]{A9D0F5}0.914 ± 0.038 & not significant \\
\textbf{Sequential Single} & \cellcolor[HTML]{C3F7C3} better & \cellcolor[HTML]{C3F7C3} better & not significant & not significant & not significant & not significant & not significant & \cellcolor[HTML]{A9D0F5}0.901 ± 0.029 \\
\bottomrule
\end{tabular}
\end{adjustbox}
\caption{AND Problem - BAcc Results with Statistical Testing: average BAcc and standard deviation over 5 folds are highlighted in blue; each row indicates statistical significance comparison with the remaining approaches: better, worse or not significant.}
\label{tab:and_bacc}
\end{table*}

\subsubsection{Interpretable Model}

Figure~\ref{fig:and_end} illustrates the resulting interpretable model for the AND problem using the end-to-end training strategy, which obtained the highest average BAcc results. With the interpretable model, one can analyze the prediction in a block-by-block fashion: the image visual explanations along with the feature values indicate a high correlation between $I_2$ and the presence of a circle, denoted by $I_{GT}$; the tabular feature $T_1$ is obtained with a piecewise symbolic expression that inversely correlates with the required feature $x_1>x_2$; the prediction $Y_{pred}$ is obtained with a symbolic expression with a binary output that is $1$ generally for very high values of $I_2$ and low values of $T_1$, excluding feature $I_1$, which is not relevant for the model. The truth table presented binarizes the feature values according to each explained extracted feature and the corresponding output. Although the model does not include the intended modality feature values, the final prediction is correct. The model found, while not appearing as anticipated, is an \emph{equivalent} model, since the induced tabular feature is just the complement of the true hidden tabular feature. Only by explaining all components in MultiFIX can we actually see this (and realize that such equivalent reasonings exist). The predictive power of the interpretable model is very similar to the black-box (DL) model, with a difference of $0.006$ in BAcc.

\subsection{XOR Problem}

\subsubsection{Problem Description}

This problem comprises the binary operation XOR between the presence of a circle~($circle$) in the image data, and the tabular feature $x_{1} > x_{2}$. The target label is thus the output of $XOR(circle,x_{1} > x_{2})$. The XOR problem reflects an extreme dependence between the two modalities since neither the image feature nor the tabular feature alone gives any information about the target label.

\subsubsection{DL Performance Results}

Figure~\ref{tab:xor_bacc} presents the performance results for the XOR Problem using single modality learning and MultiFIX with different training strategies. 
Statistical testing highlights that all multimodal approaches significantly outperform single modality approaches. Although, similarly to the AND problem, the performance values between multimodal approaches are mostly similar from a statistical perspective, their spread is larger, with the hybrid single training strategy leading to the highest average BAcc values.

\begin{table*}[]
\centering
\begin{adjustbox}{max width=0.8\textwidth}
\begin{tabular}{lcccccccc}
\toprule
 & \textbf{Image Only} & \textbf{Tabular Only} & \textbf{End-to-End} & \textbf{Hybrid AE} & \textbf{Hybrid Single} & \textbf{Sequential AE} & \textbf{Sequential AE Defreeze} & \textbf{Sequential Single} \\
\midrule
\textbf{Image Only} & \cellcolor[HTML]{A9D0F5}0.502 ± 0.027 & not significant & \cellcolor[HTML]{F6B7B7} worse & \cellcolor[HTML]{F6B7B7} worse & \cellcolor[HTML]{F6B7B7} worse & \cellcolor[HTML]{F6B7B7} worse & \cellcolor[HTML]{F6B7B7} worse & \cellcolor[HTML]{F6B7B7} worse \\
\textbf{Tabular Only} & not significant & \cellcolor[HTML]{A9D0F5}0.552 ± 0.020 & \cellcolor[HTML]{F6B7B7} worse & \cellcolor[HTML]{F6B7B7} worse & \cellcolor[HTML]{F6B7B7} worse & \cellcolor[HTML]{F6B7B7} worse & \cellcolor[HTML]{F6B7B7} worse & \cellcolor[HTML]{F6B7B7} worse \\
\textbf{End-to-End} & \cellcolor[HTML]{C3F7C3} better & \cellcolor[HTML]{C3F7C3} better & \cellcolor[HTML]{A9D0F5}0.899 ± 0.023 & not significant & not significant & not significant & not significant & not significant \\
\textbf{Hybrid AE} & \cellcolor[HTML]{C3F7C3} better & \cellcolor[HTML]{C3F7C3} better & not significant & \cellcolor[HTML]{A9D0F5}0.902 ± 0.020 & not significant & not significant & not significant & not significant \\
\textbf{Hybrid Single} & \cellcolor[HTML]{C3F7C3} better & \cellcolor[HTML]{C3F7C3} better & not significant & not significant & \cellcolor[HTML]{A9D0F5}0.918 ± 0.017 & not significant & not significant & not significant \\
\textbf{Sequential AE} & \cellcolor[HTML]{C3F7C3} better & \cellcolor[HTML]{C3F7C3} better & not significant & not significant & not significant & \cellcolor[HTML]{A9D0F5}0.812 ± 0.031 & not significant & not significant \\
\textbf{Sequential AE Defreeze} & \cellcolor[HTML]{C3F7C3} better & \cellcolor[HTML]{C3F7C3} better & not significant & not significant & not significant & not significant & \cellcolor[HTML]{A9D0F5}0.894 ± 0.046 & not significant \\
\textbf{Sequential Single} & \cellcolor[HTML]{C3F7C3} better & \cellcolor[HTML]{C3F7C3} better & not significant & not significant & not significant & not significant & not significant & \cellcolor[HTML]{A9D0F5}0.766 ± 0.041 \\
\bottomrule
\end{tabular}
\end{adjustbox}
\caption{XOR Problem - BAcc Results with Statistical Testing: average BAcc and standard deviation over 5 folds are highlighted in blue; each row indicates statistical significance comparison with the remaining approaches: better, worse or not significant.}
\label{tab:xor_bacc}
\end{table*}

\subsubsection{Interpretable Model}
\begin{figure*}[]
    \centering
    \subfigure[AND Problem End-to-End Training]{
        \includegraphics[width=0.45\textwidth]{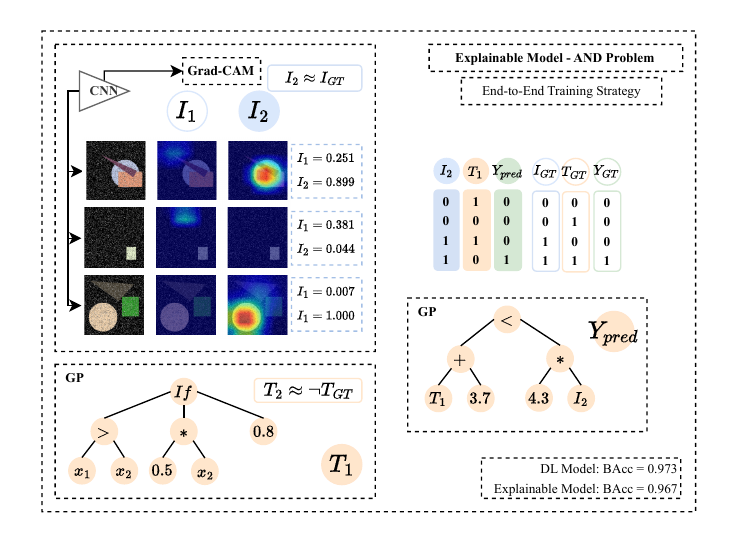}
        \label{fig:and_end}
    }
    \hspace{-7mm}
    \subfigure[XOR Problem Hybrid Single Training]{
        \includegraphics[width=0.45\textwidth]{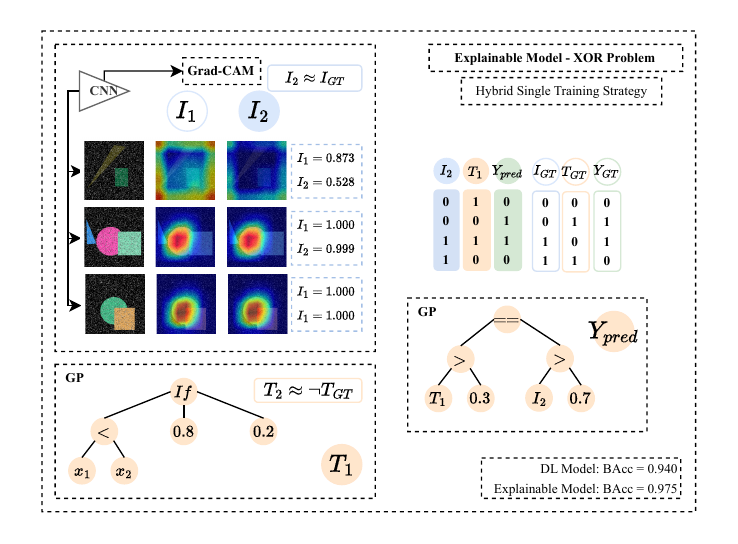}
        \label{fig:xor_hyb_single}
    }
    \caption{Interpretable Models: Grad-CAM heatmaps explain the image input contributions for each extracted feature. GP-GOMEA symbolic expressions explain the tabular features and the fusion of both modalities to make the prediction. Learned features and predictions are compared to their Ground Truth~(GT) counterparts.}
    \label{fig:xai_results}
\end{figure*}

Figure~\ref{fig:xor_hyb_single} illustrates the interpretable model for the XOR problem using the hybrid single training strategy. The image extracted features and respective explanations indicate a high correlation between $I_2$ and the presence of a circle~($I_{GT}$); the tabular feature $T_1$ is obtained with a piecewise symbolic expression inversely correlated with $x_1>x_2$ ($T_{GT}$); $Y_{pred}$ is obtained with a symbolic expression that outputs $1$ if the binarization of each useful feature using different thresholds is equal, and $0$ otherwise. The truth table presented binarizes the feature values according to each evolved threshold ($0.3$ for the tabular feature and $0.7$ for the image feature) and the corresponding prediction. Again, the complete model is correct, but the intermediate features are inverted, which leads to an equivalent model, as we can now see with MultiFIX. The predictive power of the interpretable model is higher than the black-box (DL) model, with an increase of $0.035$ in BAcc.

\subsection{Multifeature Problem}

\subsubsection{Problem Description}

In this problem, complexity is increased by scaling the number of required features per modality. It entails the following relationship:
\begin{equation}
    OR(AND(circle,x_1>x_2),AND(!triangle,x_3>x_4))
\end{equation}
Predictions of whether or not a circle and whether or not a triangle are in the image are now needed as separate features. For the tabular data, two features are also required: $x_1>x_2$ and $x_3>x_4$. The Multifeature Problem reflects a pairwise dependence between features from each modality. While each AND operation is partially dependent on both modalities, the real added difficulty of this problem comes from the need to learn each intermediate feature individually.

\subsubsection{DL Performance Results}

Figure~\ref{tab:multifeature_bacc} presents the performance results for the Multifeature Problem using single modality learning and MultiFIX with different training strategies. Average BAcc values and respective standard deviations show worse performance with single modality approaches and better performance for end-to-end and hybrid training approaches. Statistical testing, however, does not show significant differences between most of the approaches, excluding the comparison between the image approach and the sequential training approach using single modality weights. This can be justified by the extremely high variance of all multimodal approaches across different folds, and the low sample size that can reduce the sensitivity of the statistical test, despite the clear separation in BAcc ranges for different approaches.

\begin{table*}[t]
\centering
\begin{adjustbox}{max width=0.8\textwidth}
\begin{tabular}{lcccccccc}
\toprule
 & \textbf{Image Only} & \textbf{Tabular Only} & \textbf{End-to-End} & \textbf{Hybrid AE} & \textbf{Hybrid Single} & \textbf{Sequential AE} & \textbf{Sequential AE Defreeze} & \textbf{Sequential Single} \\
\midrule
\textbf{Image Only} & \cellcolor[HTML]{A9D0F5}0.655 ± 0.006 & not significant & \cellcolor[HTML]{F6B7B7} not significant & \cellcolor[HTML]{F6B7B7} not significant & \cellcolor[HTML]{F6B7B7} not significant & \cellcolor[HTML]{F6B7B7} not significant & \cellcolor[HTML]{F6B7B7} not significant & \cellcolor[HTML]{F6B7B7} worse \\
\textbf{Tabular Only} & not significant & \cellcolor[HTML]{A9D0F5}0.674 ± 0.028 & \cellcolor[HTML]{F6B7B7} not significant & \cellcolor[HTML]{F6B7B7} not significant & \cellcolor[HTML]{F6B7B7} not significant & \cellcolor[HTML]{F6B7B7} not significant & \cellcolor[HTML]{F6B7B7} not significant & \cellcolor[HTML]{F6B7B7} not significant \\
\textbf{End-to-End} & \cellcolor[HTML]{C3F7C3} not significant & \cellcolor[HTML]{C3F7C3} not significant & \cellcolor[HTML]{A9D0F5}0.798 ± 0.044 & not significant & not significant & \cellcolor[HTML]{C3F7C3} not significant & not significant & not significant \\
\textbf{Hybrid AE} & \cellcolor[HTML]{C3F7C3} not significant & \cellcolor[HTML]{C3F7C3} not significant & not significant & \cellcolor[HTML]{A9D0F5}0.799 ± 0.054 & not significant & not significant & not significant & not significant \\
\textbf{Hybrid Single} & \cellcolor[HTML]{C3F7C3} not significant & \cellcolor[HTML]{C3F7C3} not significant & not significant & not significant & \cellcolor[HTML]{A9D0F5}0.798 ± 0.058 & not significant & not significant & not significant \\
\textbf{Sequential AE} & \cellcolor[HTML]{C3F7C3} not significant & \cellcolor[HTML]{C3F7C3} not significant & \cellcolor[HTML]{F6B7B7} not significant & not significant & not significant & \cellcolor[HTML]{A9D0F5}0.703 ± 0.046 & not significant & not significant \\
\textbf{Sequential AE Defreeze} & \cellcolor[HTML]{C3F7C3} not significant & \cellcolor[HTML]{C3F7C3} not significant & not significant & not significant & not significant & not significant & \cellcolor[HTML]{A9D0F5}0.766 ± 0.077 & not significant \\
\textbf{Sequential Single} & \cellcolor[HTML]{C3F7C3} better & \cellcolor[HTML]{C3F7C3} not significant & not significant & not significant & not significant & not significant & not significant & \cellcolor[HTML]{A9D0F5}0.743 ± 0.018 \\
\bottomrule
\end{tabular}
\end{adjustbox}
\caption{Multifeature Problem - BAcc Results with Statistical Testing: average BAcc and standard deviation over 5 folds are highlighted in blue; each row indicates statistical significance comparison with the remaining approaches: better, worse or not significant.}
\label{tab:multifeature_bacc}
\end{table*}

\subsubsection{Interpretable Model}

Figure~\ref{fig:andor_hyb_ae} illustrates the resulting interpretable model for the Multifeature problem using the hybrid training strategy with encoder weights for the pre-trained image feature engineering block, which obtained the highest average performance over the remaining approaches. The inherent added complexity of this problem is reflected in the resulting interpretable model. In the image feature engineering block, complex features were learned. Naively, one would hope that each extracted feature would exclusively relate to one of the shapes of interest (circle and triangle). However, the resulting intermediate features, although comprising relevant information, are not as simple to analyze. Plotting samples with different image characteristics in a 2D space with $I_1$ and $I_2$ on the axes revealed that lower values of $I_1$ are correlated with the presence of a triangle \emph{and} the absence of a circle in the image; very high values of $I_2$ (approximately higher than $0.8$) are correlated with the absence of \emph{both} shapes in the image; samples in which $I_1>I_2$ are correlated with the presence of a circle in the image; when a circle is present, the presence or absence of a triangle is not well distinguished (which indicates samples for which the model fails). The reason for this happening is that the intermediate image features are set to be real-valued, making it possible to map multiple features that are essentially binary to subranges of one real-valued feature. This analysis is corroborated with the analyzed plot in the Supplementary Material. The fusion symbolic expression with the highest performance is a tree with depth three, while for the problems so far, a tree with depth two sufficed. This increases complexity and arguably decreases interpretability. Despite being still transparent and readable, the obtained expression is not easy to interpret. The condition is related to whether there is a circle in the image and the then-branch is related to the relation between $x_1$ and $x_2$. The else-branch is related to the relation between $x_3$ and $x_4$, as well as whether there is a triangle in the image, which signals correct logic in terms of features involved. The tabular feature engineering block reveals symbolic expressions that accurately correlate to the two required features: $T_1$ is inversely correlated with $x_1>x_2$, presenting values larger than $0.4$ when $x_1<x_2$, and values smaller or equal to $0.4$ otherwise; $T_2$ and $T_3$ are correlated and inversely correlated (respectively) with $x_3>x_4$, using different evolved binarization thresholds. Lastly, the predictive power of the interpretable model is higher than the black-box (DL) model, with an increase of $0.020$ in BAcc.

\subsection{Multiclass Problem}

\subsubsection{Problem Description}

The last problem has a multiclass target with four possible classes rather than a binary target. The classes relate to combinations of the intermediate features of each modality (presence of circle in an image, and $x_1>x_2$), as demonstrated in Table~\ref{tab:multiclass_table}. Despite each modality being correlated with the endpoint, optimal predictions need joint information from both inputs.
\vspace{-4mm}
\begin{table}[h]
    \centering
    \begin{tabular}{c c|c}
         $ft_{circle}$ & $ft_{x_1>x_2}$ & $Y_{multiclass}$\\ \hline
         0 & 0 & 0 \\
         0 & 1 & 1 \\
         1 & 0 & 2 \\
         1 & 1 & 3
    \end{tabular}
    \caption{Multiclass Problem using $ft_{circle}$ and  $ft_{x_1>x_2}$.}
    \label{tab:multiclass_table}
\end{table}

\vspace{-8mm}
\subsubsection{DL Performance Results}

In figure~\ref{tab:multiclass_bacc} the performance results for the Multiclass Problem using single modality learning and MultiFIX with different training strategies are shown. Statistical testing highlights that all multimodal approaches significantly outperform single modality approaches, excluding the sequential training approach using AE weights in the image block. Similarly to the AND and XOR problems, the different multimodal approaches are close in performance from a statistical significance perspective, although the hybrid single training strategy achieves higher average BAcc values.

\begin{table*}[]
\centering
\begin{adjustbox}{max width=0.8\textwidth}
\begin{tabular}{lcccccccc}
\toprule
 & \textbf{Image Only} & \textbf{Tabular Only} & \textbf{End-to-End} & \textbf{Hybrid AE} & \textbf{Hybrid Single} & \textbf{Sequential AE} & \textbf{Sequential AE Defreeze} & \textbf{Sequential Single} \\
\midrule
\textbf{Image Only} & \cellcolor[HTML]{A9D0F5}0.485 ± 0.004 & not significant & \cellcolor[HTML]{F6B7B7} worse & \cellcolor[HTML]{F6B7B7} worse & \cellcolor[HTML]{F6B7B7} worse & \cellcolor[HTML]{F6B7B7} not significant & \cellcolor[HTML]{F6B7B7} worse & \cellcolor[HTML]{F6B7B7} worse \\
\textbf{Tabular Only} & not significant & \cellcolor[HTML]{A9D0F5}0.453 ± 0.014 & \cellcolor[HTML]{F6B7B7} worse & \cellcolor[HTML]{F6B7B7} worse & \cellcolor[HTML]{F6B7B7} worse & \cellcolor[HTML]{F6B7B7} worse & \cellcolor[HTML]{F6B7B7} worse & \cellcolor[HTML]{F6B7B7} worse \\
\textbf{End-to-End} & \cellcolor[HTML]{C3F7C3} better & \cellcolor[HTML]{C3F7C3} better & \cellcolor[HTML]{A9D0F5}0.823 ± 0.038 & not significant & not significant & \cellcolor[HTML]{C3F7C3} not significant & not significant & not significant \\
\textbf{Hybrid AE} & \cellcolor[HTML]{C3F7C3} better & \cellcolor[HTML]{C3F7C3} better & not significant & \cellcolor[HTML]{A9D0F5}0.858 ± 0.036 & not significant & not significant & not significant & not significant \\
\textbf{Hybrid Single} & \cellcolor[HTML]{C3F7C3} better & \cellcolor[HTML]{C3F7C3} better & not significant & not significant & \cellcolor[HTML]{A9D0F5}0.919 ± 0.007 & not significant & not significant & not significant \\
\textbf{Sequential AE} & \cellcolor[HTML]{C3F7C3} not significant & \cellcolor[HTML]{C3F7C3} better & \cellcolor[HTML]{F6B7B7} not significant & not significant & not significant & \cellcolor[HTML]{A9D0F5}0.691 ± 0.061 & not significant & not significant \\
\textbf{Sequential AE Defreeze} & \cellcolor[HTML]{C3F7C3} better & \cellcolor[HTML]{C3F7C3} better & not significant & not significant & not significant & not significant & \cellcolor[HTML]{A9D0F5}0.849 ± 0.053 & not significant \\
\textbf{Sequential Single} & \cellcolor[HTML]{C3F7C3} better & \cellcolor[HTML]{C3F7C3} better & not significant & not significant & not significant & not significant & not significant & \cellcolor[HTML]{A9D0F5}0.742 ± 0.055 \\
\bottomrule
\end{tabular}
\end{adjustbox}
\caption{Multiclass Problem - BAcc Results with Statistical Testing: average BAcc and standard deviation over 5 folds are highlighted in blue; each row indicates statistical significance comparison with the remaining approaches: better, worse or not significant.}
\label{tab:multiclass_bacc}
\end{table*}

\subsubsection{Interpretable Model}

\begin{figure*}[]
    \centering
    \subfigure[Multifeature Problem Hybrid  AE Training]{
        \includegraphics[width=0.42\textwidth]{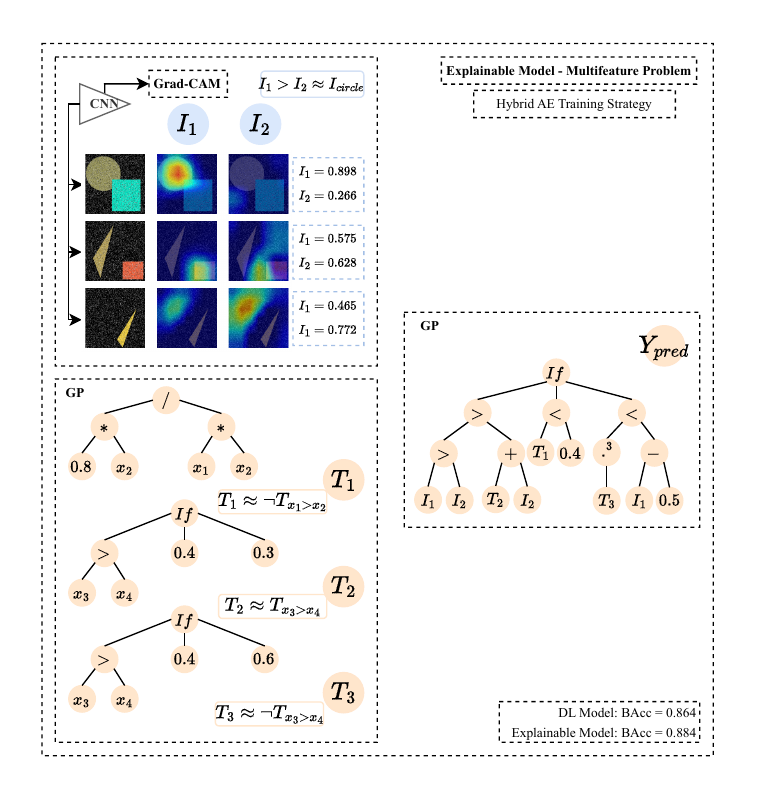}
        \label{fig:andor_hyb_ae}
    }
    \hspace{-7mm}
    \subfigure[Multiclass Problem Hybrid Single Training]{
        \includegraphics[width=0.48\textwidth]{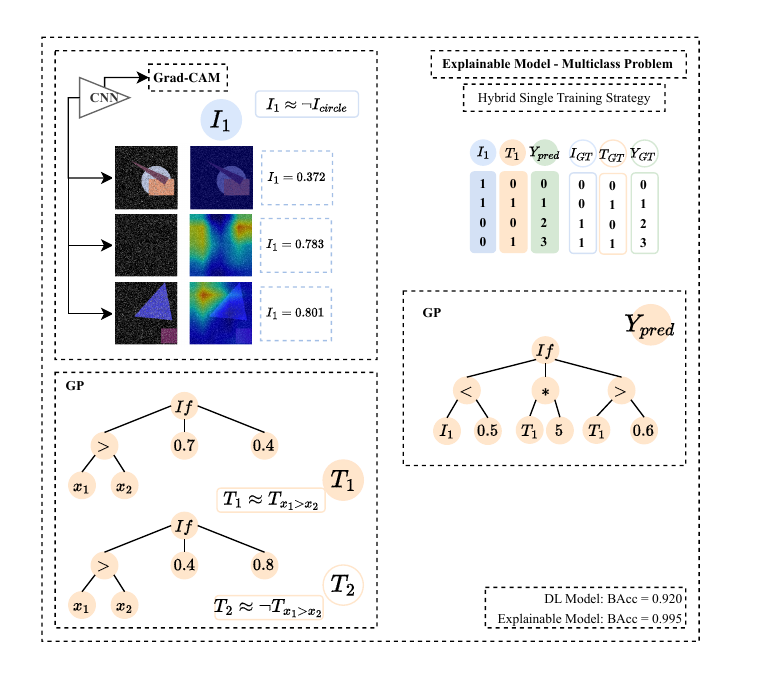}
        \label{fig:multiclass_hyb_single}
    }

    \caption{Interpretable Models: Grad-CAM heatmaps explain the image input contributions for each extracted feature. GP-GOMEA symbolic expressions explain the tabular features and the fusion of both modalities to make the prediction. Learned features and predictions are compared to their GT counterparts.}
    \label{fig:xai_results}
\end{figure*}

Figure~\ref{fig:multiclass_hyb_single} illustrates the resulting interpretable model for the Multiclass problem using the hybrid single training strategy. The image visual explanations along with the feature values indicate a high correlation between $I_1$ and the absence of a circle, which can be inferred from a very low feature value and a low-contribution heatmap when a circle is present; the tabular feature $T_1$ is obtained with a piecewise symbolic expression that directly correlates with the engineered feature $x_1>x_2$; the prediction $Y_{pred}$ is calculated with a piecewise symbolic expression with the condition $I_1<0.5$ (is a circle present in the image), and with two possible subtrees that use $T_1$ to assign the values $2$ or $3$, if a circle is in the image, and $0$ or $1$, if a circle is absent in the image. The predictive power of the interpretable model is substantially higher than the black-box (DL) model, with an increase of $0.075$ in BAcc.

\section{Discussion}
\label{sec:discussion}

MultiFIX is a unique approach to explainable multimodal learning that enforces a small number of features to be learned from different modalities. The strengths of DL and GP are leveraged to learn interpretable models that integrate symbolic expressions for both interpretable tabular feature extraction and interpretable fusion of different modalities, and image explanation techniques such as Grad-CAM. Our results showcase the potential for developing models that are interpretable by design without sacrificing performance, specifically in problems that include tabular data or combine heterogeneous data modalities. This work contributes to the growing need for explainable multimodal methods in high-stakes domains, such as healthcare.

In comparison to post-hoc methods like SHAP, GP-GOMEA directly generates interpretable models without additional approximations, reducing the risk of misleading explanations. Furthermore, the incorporation of interpretability at the feature level enables directly understanding which variables are relevant for the task at hand. Lastly, our experiments indicate that end-to-end and hybrid training achieve better performance than sequential training approaches. However, statistical testing indicates that most of the studied training approaches obtain similar performance. Thus, the choice of training strategy should rely on the specifications of the problem. For instance, one may have a powerful feature engineering block for one of the modalities that can be used either as pre-training weights or as a sequential block. Additionally, in many multimodal problems, there are different amounts of samples per modality, which usually translates in removing samples from the oversampled modality. Using hybrid and sequential training, the total amount of samples can be leveraged to learn more accurate feature engineering blocks.

While providing a unique step toward interpretable multimodal learning, the way we have trained and used MutiFIX so far has clear limitations as well. Firstly, we have here considered a limited set of problems with a limited number of modalities. While this is important to be able to fundamentally study the possibilities and limitations of MultiFIX when the ground truth is known, the added value offered by MultiFIX must be corroborated on real-world datasets, which we intend to do in the near future.

When more than one complex feature is needed, as demonstrated in the Multifeature Problem, the complexity of the intermediate image features can hinder interpretability. Moreover, requiring more features in the bottleneck creates possibilities to obtain different, but equivalent models, making it not always easy to interpret what the model is doing, as at first it may be counterintuitive. Forcing the image block to engineer simpler features can reduce the overall complexity and increase interpretability. More generally, the penalization of complexity as an additional objective is likely advantageous.
For the discovery of symbolic expressions, using a multi-objective variant of GP-GOMEA could help to find models of different sizes, some of which are more easily interpretable in their own right, or may provide hints as to what slightly larger (and potentially more accurate) expressions are capable of modeling.

The use of Grad-CAM to explain image feature extraction highlights regions used by the DL model that contribute to the engineered feature. However, the post-hoc nature of Grad-CAM limits the symbolic meaning that can be associated with each feature. Having inherently interpretable image blocks would be highly beneficial from an interpretability perspective. Lastly, the interpretable models generated by MultiFIX, although readable and transparent, can further benefit from user-friendly presentations that can enhance overall interpretability. This includes further visual explanations, approximations, and simplifications for the user, as well as ways to perform interactive and sample-based analyses.

\section{Conclusion}
\label{sec:conclusion}

In this paper, we present the first comprehensive experimental work on MultiFIX, a novel multimodal pipeline to obtain multimodal, interpretable models. The uniqueness of MultiFIX is reflected in its interpretability-focused design by forcing a limited number of features per modality to be automatically engineered and subsequently used to make predictions. DL can be used to perform feature engineering, whereas GP-GOMEA can be used to evolve interpretable symbolic expressions for tabular engineered features and for the final fusion. Modality-specific post-hoc explainability techniques can be used, such as Grad-CAM for images to explain the overall model in a component-wise fashion.

Considering the different multimodal training strategies that we have studied, there seems to be no statistically significant difference in performance for the created benchmark problems. The choice of the most suitable training strategy is up to the specifications of the task at hand. In general, our results demonstrated that MultiFIX can accurately capture multimodal relationships and that learned models have high potential for interpretability.

Despite these advancements, MultiFIX needs further improvements, predominantly interpretability enhancements that minimize complexity and promote inherently interpretable methods for image features, in combination with additional interactive visualization tools to intuitively clarify the mechanics of the learned models. Furthermore, we aim to evaluate our pipeline on real-world datasets that involve more modalities as well as more complex intermodal relationships.

In conclusion, we believe that we have demonstrated the feasibility and potential to perform interpretable multimodal learning by leveraging a unique feature-inducing architecture combined with inherently interpretable methods across heterogeneous data types with MultiFIX.

\begin{acks}
This research is part of the "Uitlegbare Kunstmatige Intelligentie" project funded by the Stichting Gieskes-Strijbis Fonds. We also thank NWO for the Small Compute grant on the Dutch National Supercomputer Snellius.
\end{acks}

\bibliographystyle{ACM-Reference-Format}

\begin{thebibliography}{27}


\ifx \showCODEN    \undefined \def \showCODEN     #1{\unskip}     \fi
\ifx \showDOI      \undefined \def \showDOI       #1{#1}\fi
\ifx \showISBNx    \undefined \def \showISBNx     #1{\unskip}     \fi
\ifx \showISBNxiii \undefined \def \showISBNxiii  #1{\unskip}     \fi
\ifx \showISSN     \undefined \def \showISSN      #1{\unskip}     \fi
\ifx \showLCCN     \undefined \def \showLCCN      #1{\unskip}     \fi
\ifx \shownote     \undefined \def \shownote      #1{#1}          \fi
\ifx \showarticletitle \undefined \def \showarticletitle #1{#1}   \fi
\ifx \showURL      \undefined \def \showURL       {\relax}        \fi
\providecommand\bibfield[2]{#2}
\providecommand\bibinfo[2]{#2}
\providecommand\natexlab[1]{#1}
\providecommand\showeprint[2][]{arXiv:#2}

\bibitem[Allgaier et~al\mbox{.}(2023)]%
        {allgaier2023grad}
\bibfield{author}{\bibinfo{person}{Johannes Allgaier}, \bibinfo{person}{Lena Mulansky}, \bibinfo{person}{Rachel Draelos}, {and} \bibinfo{person}{Rüdiger Pryss}.} \bibinfo{year}{2023}\natexlab{}.
\newblock \showarticletitle{How does the model make predictions? A systematic literature review on the explainability power of machine learning in healthcare}.
\newblock \bibinfo{journal}{\emph{Artificial Intelligence in Medicine}}  \bibinfo{volume}{143} (\bibinfo{date}{09} \bibinfo{year}{2023}), \bibinfo{pages}{102616}.
\newblock
\urldef\tempurl%
\url{https://doi.org/10.1016/j.artmed.2023.102616}
\showDOI{\tempurl}


\bibitem[Arora et~al\mbox{.}(2023)]%
        {arora2023introduction}
\bibfield{author}{\bibinfo{person}{Nitin Arora}, \bibinfo{person}{Anupam Singh}, \bibinfo{person}{Vivek Shahare}, {and} \bibinfo{person}{Goutam Datta}.} \bibinfo{year}{2023}\natexlab{}.
\newblock \showarticletitle{Introduction to Big Data Analytics}.
\newblock In \bibinfo{booktitle}{\emph{Towards the Integration of IoT, Cloud and Big Data: Services, Applications and Standards}}. \bibinfo{publisher}{Springer}, \bibinfo{pages}{1--18}.
\newblock


\bibitem[Bacardit et~al\mbox{.}(2022)]%
        {bacardit2022intersection}
\bibfield{author}{\bibinfo{person}{Jaume Bacardit}, \bibinfo{person}{Alexander~EI Brownlee}, \bibinfo{person}{Stefano Cagnoni}, \bibinfo{person}{Giovanni Iacca}, \bibinfo{person}{John McCall}, {and} \bibinfo{person}{David Walker}.} \bibinfo{year}{2022}\natexlab{}.
\newblock \showarticletitle{The intersection of evolutionary computation and explainable AI}. In \bibinfo{booktitle}{\emph{Proceedings of the Genetic and Evolutionary Computation conference companion}}. \bibinfo{pages}{1757--1762}.
\newblock


\bibitem[Chen et~al\mbox{.}(2022)]%
        {chen2022pan}
\bibfield{author}{\bibinfo{person}{Richard~J Chen}, \bibinfo{person}{Ming~Y Lu}, \bibinfo{person}{Drew~FK Williamson}, \bibinfo{person}{Tiffany~Y Chen}, \bibinfo{person}{Jana Lipkova}, \bibinfo{person}{Zahra Noor}, \bibinfo{person}{Muhammad Shaban}, \bibinfo{person}{Maha Shady}, \bibinfo{person}{Mane Williams}, \bibinfo{person}{Bumjin Joo}, {et~al\mbox{.}}} \bibinfo{year}{2022}\natexlab{}.
\newblock \showarticletitle{Pan-cancer integrative histology-genomic analysis via multimodal deep learning}.
\newblock \bibinfo{journal}{\emph{Cancer Cell}} \bibinfo{volume}{40}, \bibinfo{number}{8} (\bibinfo{year}{2022}), \bibinfo{pages}{865--878}.
\newblock


\bibitem[Evans et~al\mbox{.}(2019)]%
        {evans2019s}
\bibfield{author}{\bibinfo{person}{Benjamin~P Evans}, \bibinfo{person}{Bing Xue}, {and} \bibinfo{person}{Mengjie Zhang}.} \bibinfo{year}{2019}\natexlab{}.
\newblock \showarticletitle{What's inside the black-box? a genetic programming method for interpreting complex machine learning models}. In \bibinfo{booktitle}{\emph{Proceedings of the genetic and evolutionary computation conference}}. \bibinfo{pages}{1012--1020}.
\newblock


\bibitem[Gildenblat and contributors(2021)]%
        {jacobgilpytorchcam}
\bibfield{author}{\bibinfo{person}{Jacob Gildenblat} {and} \bibinfo{person}{contributors}.} \bibinfo{year}{2021}\natexlab{}.
\newblock \bibinfo{title}{PyTorch library for CAM methods}.
\newblock \bibinfo{howpublished}{\url{https://github.com/jacobgil/pytorch-grad-cam}}.
\newblock


\bibitem[Guarrasi et~al\mbox{.}(2024)]%
        {guarrasi2024systematic}
\bibfield{author}{\bibinfo{person}{Valerio Guarrasi}, \bibinfo{person}{Fatih Aksu}, \bibinfo{person}{Camillo~Maria Caruso}, \bibinfo{person}{Francesco Di~Feola}, \bibinfo{person}{Aurora Rofena}, \bibinfo{person}{Filippo Ruffini}, {and} \bibinfo{person}{Paolo Soda}.} \bibinfo{year}{2024}\natexlab{}.
\newblock \showarticletitle{A Systematic Review of Intermediate Fusion in Multimodal Deep Learning for Biomedical Applications}.
\newblock \bibinfo{journal}{\emph{arXiv preprint arXiv:2408.02686}} (\bibinfo{year}{2024}).
\newblock


\bibitem[He et~al\mbox{.}(2016)]%
        {resnet18}
\bibfield{author}{\bibinfo{person}{Kaiming He}, \bibinfo{person}{Xiangyu Zhang}, \bibinfo{person}{Shaoqing Ren}, {and} \bibinfo{person}{Jian Sun}.} \bibinfo{year}{2016}\natexlab{}.
\newblock \showarticletitle{Deep Residual Learning for Image Recognition}. In \bibinfo{booktitle}{\emph{2016 IEEE Conference on Computer Vision and Pattern Recognition (CVPR)}}. \bibinfo{pages}{770--778}.
\newblock
\urldef\tempurl%
\url{https://doi.org/10.1109/CVPR.2016.90}
\showDOI{\tempurl}


\bibitem[Huang et~al\mbox{.}(2020)]%
        {huang2020fusion}
\bibfield{author}{\bibinfo{person}{Shih-Cheng Huang}, \bibinfo{person}{Anuj Pareek}, \bibinfo{person}{Saeed Seyyedi}, \bibinfo{person}{Imon Banerjee}, {and} \bibinfo{person}{Matthew~P Lungren}.} \bibinfo{year}{2020}\natexlab{}.
\newblock \showarticletitle{Fusion of medical imaging and electronic health records using deep learning: a systematic review and implementation guidelines}.
\newblock \bibinfo{journal}{\emph{NPJ digital medicine}} \bibinfo{volume}{3}, \bibinfo{number}{1} (\bibinfo{year}{2020}), \bibinfo{pages}{136}.
\newblock


\bibitem[Joshi et~al\mbox{.}(2021)]%
        {joshi2021review}
\bibfield{author}{\bibinfo{person}{Gargi Joshi}, \bibinfo{person}{Rahee Walambe}, {and} \bibinfo{person}{Ketan Kotecha}.} \bibinfo{year}{2021}\natexlab{}.
\newblock \showarticletitle{A review on explainability in multimodal deep neural nets}.
\newblock \bibinfo{journal}{\emph{IEEE Access}}  \bibinfo{volume}{9} (\bibinfo{year}{2021}), \bibinfo{pages}{59800--59821}.
\newblock


\bibitem[Kingma and Ba(2015)]%
        {kingma2015adam}
\bibfield{author}{\bibinfo{person}{Diederik~P Kingma} {and} \bibinfo{person}{Jimmy Ba}.} \bibinfo{year}{2015}\natexlab{}.
\newblock \showarticletitle{Adam: A Method for Stochastic Optimization}.
\newblock \bibinfo{journal}{\emph{International Conference on Learning Representations (ICLR)}} (\bibinfo{year}{2015}).
\newblock


\bibitem[Kline et~al\mbox{.}(2022)]%
        {kline2022multimodal}
\bibfield{author}{\bibinfo{person}{Adrienne Kline}, \bibinfo{person}{Hanyin Wang}, \bibinfo{person}{Yikuan Li}, \bibinfo{person}{Saya Dennis}, \bibinfo{person}{Meghan Hutch}, \bibinfo{person}{Zhenxing Xu}, \bibinfo{person}{Fei Wang}, \bibinfo{person}{Feixiong Cheng}, {and} \bibinfo{person}{Yuan Luo}.} \bibinfo{year}{2022}\natexlab{}.
\newblock \showarticletitle{Multimodal machine learning in precision health: A scoping review}.
\newblock \bibinfo{journal}{\emph{npj Digital Medicine}} \bibinfo{volume}{5}, \bibinfo{number}{1} (\bibinfo{year}{2022}), \bibinfo{pages}{171}.
\newblock


\bibitem[La~Cava et~al\mbox{.}(2021)]%
        {srbench}
\bibfield{author}{\bibinfo{person}{William La~Cava}, \bibinfo{person}{Patryk Orzechowski}, \bibinfo{person}{Bogdan Burlacu}, \bibinfo{person}{Fabricio de Franca}, \bibinfo{person}{Marco Virgolin}, \bibinfo{person}{Ying Jin}, \bibinfo{person}{Michael Kommenda}, {and} \bibinfo{person}{Jason Moore}.} \bibinfo{year}{2021}\natexlab{}.
\newblock \showarticletitle{Contemporary Symbolic Regression Methods and their Relative Performance}. In \bibinfo{booktitle}{\emph{Proceedings of the Neural Information Processing Systems Track on Datasets and Benchmarks}}, \bibfield{editor}{\bibinfo{person}{J.~Vanschoren} {and} \bibinfo{person}{S.~Yeung}} (Eds.), Vol.~\bibinfo{volume}{1}. \bibinfo{publisher}{Curran}.
\newblock


\bibitem[Malafaia et~al\mbox{.}(2024)]%
        {malafaia2024multifix}
\bibfield{author}{\bibinfo{person}{Mafalda Malafaia}, \bibinfo{person}{Thalea Schlender}, \bibinfo{person}{Peter~AN Bosman}, {and} \bibinfo{person}{Tanja Alderliesten}.} \bibinfo{year}{2024}\natexlab{}.
\newblock \showarticletitle{MultiFIX: An XAI-friendly feature inducing approach to building models from multimodal data}.
\newblock \bibinfo{journal}{\emph{arXiv preprint arXiv:2402.12183}} (\bibinfo{year}{2024}).
\newblock


\bibitem[Rahate et~al\mbox{.}(2022)]%
        {rahate2022multimodal}
\bibfield{author}{\bibinfo{person}{Anil Rahate}, \bibinfo{person}{Rahee Walambe}, \bibinfo{person}{Sheela Ramanna}, {and} \bibinfo{person}{Ketan Kotecha}.} \bibinfo{year}{2022}\natexlab{}.
\newblock \showarticletitle{Multimodal co-learning: Challenges, applications with datasets, recent advances and future directions}.
\newblock \bibinfo{journal}{\emph{Information Fusion}}  \bibinfo{volume}{81} (\bibinfo{year}{2022}), \bibinfo{pages}{203--239}.
\newblock


\bibitem[Rudin(2019)]%
        {rudin2019stop}
\bibfield{author}{\bibinfo{person}{Cynthia Rudin}.} \bibinfo{year}{2019}\natexlab{}.
\newblock \showarticletitle{Stop explaining black box machine learning models for high stakes decisions and use interpretable models instead}.
\newblock \bibinfo{journal}{\emph{Nature machine intelligence}} \bibinfo{volume}{1}, \bibinfo{number}{5} (\bibinfo{year}{2019}), \bibinfo{pages}{206--215}.
\newblock


\bibitem[Rudin et~al\mbox{.}(2022)]%
        {rudin2022interpretable}
\bibfield{author}{\bibinfo{person}{Cynthia Rudin}, \bibinfo{person}{Chaofan Chen}, \bibinfo{person}{Zhi Chen}, \bibinfo{person}{Haiyang Huang}, \bibinfo{person}{Lesia Semenova}, {and} \bibinfo{person}{Chudi Zhong}.} \bibinfo{year}{2022}\natexlab{}.
\newblock \showarticletitle{Interpretable machine learning: Fundamental principles and 10 grand challenges}.
\newblock \bibinfo{journal}{\emph{Statistic Surveys}}  \bibinfo{volume}{16} (\bibinfo{year}{2022}), \bibinfo{pages}{1--85}.
\newblock


\bibitem[Schlender et~al\mbox{.}(2024)]%
        {schlender2024improving}
\bibfield{author}{\bibinfo{person}{Thalea Schlender}, \bibinfo{person}{Mafalda Malafaia}, \bibinfo{person}{Tanja Alderliesten}, {and} \bibinfo{person}{Peter Bosman}.} \bibinfo{year}{2024}\natexlab{}.
\newblock \showarticletitle{Improving the efficiency of GP-GOMEA for higher-arity operators}. In \bibinfo{booktitle}{\emph{Proceedings of the Genetic and Evolutionary Computation Conference}}. \bibinfo{pages}{971--979}.
\newblock


\bibitem[Schouten et~al\mbox{.}(2024)]%
        {schouten2024navigating}
\bibfield{author}{\bibinfo{person}{Daan Schouten}, \bibinfo{person}{Giulia Nicoletti}, \bibinfo{person}{Bas Dille}, \bibinfo{person}{Catherine Chia}, \bibinfo{person}{Pierpaolo Vendittelli}, \bibinfo{person}{Megan Schuurmans}, \bibinfo{person}{Geert Litjens}, {and} \bibinfo{person}{Nadieh Khalili}.} \bibinfo{year}{2024}\natexlab{}.
\newblock \showarticletitle{Navigating the landscape of multimodal AI in medicine: a scoping review on technical challenges and clinical applications}.
\newblock \bibinfo{journal}{\emph{arXiv preprint arXiv:2411.03782}} (\bibinfo{year}{2024}).
\newblock


\bibitem[Selvaraju et~al\mbox{.}(2019)]%
        {grad_cam19}
\bibfield{author}{\bibinfo{person}{Ramprasaath~R. Selvaraju}, \bibinfo{person}{Michael Cogswell}, \bibinfo{person}{Abhishek Das}, \bibinfo{person}{Ramakrishna Vedantam}, \bibinfo{person}{Devi Parikh}, {and} \bibinfo{person}{Dhruv Batra}.} \bibinfo{year}{2019}\natexlab{}.
\newblock \showarticletitle{Grad-CAM: Visual Explanations from Deep Networks via Gradient-Based Localization}.
\newblock \bibinfo{journal}{\emph{International Journal of Computer Vision}} \bibinfo{volume}{128}, \bibinfo{number}{2} (\bibinfo{year}{2019}), \bibinfo{pages}{336–359}.
\newblock
\showISSN{1573-1405}
\urldef\tempurl%
\url{https://doi.org/10.1007/s11263-019-01228-7}
\showDOI{\tempurl}


\bibitem[Sleeman et~al\mbox{.}(2022)]%
        {sleeman2022multimodal}
\bibfield{author}{\bibinfo{person}{William~C Sleeman}, \bibinfo{person}{Rishabh Kapoor}, {and} \bibinfo{person}{Preetam Ghosh}.} \bibinfo{year}{2022}\natexlab{}.
\newblock \showarticletitle{Multimodal classification: Current landscape, taxonomy and future directions}.
\newblock \bibinfo{journal}{\emph{Comput. Surveys}} \bibinfo{volume}{55}, \bibinfo{number}{7} (\bibinfo{year}{2022}), \bibinfo{pages}{1--31}.
\newblock


\bibitem[Stahlschmidt et~al\mbox{.}(2022)]%
        {stahlschmidt2022multimodal}
\bibfield{author}{\bibinfo{person}{S{\"o}ren~Richard Stahlschmidt}, \bibinfo{person}{Benjamin Ulfenborg}, {and} \bibinfo{person}{Jane Synnergren}.} \bibinfo{year}{2022}\natexlab{}.
\newblock \showarticletitle{Multimodal deep learning for biomedical data fusion: a review}.
\newblock \bibinfo{journal}{\emph{Briefings in Bioinformatics}} \bibinfo{volume}{23}, \bibinfo{number}{2} (\bibinfo{year}{2022}), \bibinfo{pages}{bbab569}.
\newblock


\bibitem[Swamy et~al\mbox{.}(2024)]%
        {swamy2024multimodn}
\bibfield{author}{\bibinfo{person}{Vinitra Swamy}, \bibinfo{person}{Malika Satayeva}, \bibinfo{person}{Jibril Frej}, \bibinfo{person}{Thierry Bossy}, \bibinfo{person}{Thijs Vogels}, \bibinfo{person}{Martin Jaggi}, \bibinfo{person}{Tanja K{\"a}ser}, {and} \bibinfo{person}{Mary-Anne Hartley}.} \bibinfo{year}{2024}\natexlab{}.
\newblock \showarticletitle{Multimodn—multimodal, multi-task, interpretable modular networks}.
\newblock \bibinfo{journal}{\emph{Advances in Neural Information Processing Systems}}  \bibinfo{volume}{36} (\bibinfo{year}{2024}).
\newblock


\bibitem[Virgolin et~al\mbox{.}(2020)]%
        {virgolin2020explaining}
\bibfield{author}{\bibinfo{person}{Marco Virgolin}, \bibinfo{person}{Tanja Alderliesten}, {and} \bibinfo{person}{Peter~AN Bosman}.} \bibinfo{year}{2020}\natexlab{}.
\newblock \showarticletitle{On explaining machine learning models by evolving crucial and compact features}.
\newblock \bibinfo{journal}{\emph{Swarm and Evolutionary Computation}}  \bibinfo{volume}{53} (\bibinfo{year}{2020}), \bibinfo{pages}{100640}.
\newblock


\bibitem[Virgolin et~al\mbox{.}(2021)]%
        {gp_gomea21}
\bibfield{author}{\bibinfo{person}{Marco Virgolin}, \bibinfo{person}{Tanja Alderliesten}, \bibinfo{person}{Cees Witteveen}, {and} \bibinfo{person}{Peter A.~N. Bosman}.} \bibinfo{year}{2021}\natexlab{}.
\newblock \showarticletitle{Improving model-based genetic programming for symbolic regression of small expressions}.
\newblock \bibinfo{journal}{\emph{Evolutionary Computation}} \bibinfo{volume}{29}, \bibinfo{number}{2} (\bibinfo{year}{2021}), \bibinfo{pages}{211--237}.
\newblock


\bibitem[Zhao et~al\mbox{.}(2024)]%
        {zhao2024deep}
\bibfield{author}{\bibinfo{person}{Fei Zhao}, \bibinfo{person}{Chengcui Zhang}, {and} \bibinfo{person}{Baocheng Geng}.} \bibinfo{year}{2024}\natexlab{}.
\newblock \showarticletitle{Deep Multimodal Data Fusion}.
\newblock \bibinfo{journal}{\emph{Comput. Surveys}} \bibinfo{volume}{56}, \bibinfo{number}{9} (\bibinfo{year}{2024}), \bibinfo{pages}{1--36}.
\newblock


\bibitem[Zhou and Hu(2023)]%
        {zhou2023evolutionary}
\bibfield{author}{\bibinfo{person}{Ryan Zhou} {and} \bibinfo{person}{Ting Hu}.} \bibinfo{year}{2023}\natexlab{}.
\newblock \showarticletitle{Evolutionary approaches to explainable machine learning}.
\newblock In \bibinfo{booktitle}{\emph{Handbook of Evolutionary Machine Learning}}. \bibinfo{publisher}{Springer}, \bibinfo{pages}{487--506}.
\newblock


\end{thebibliography}

\appendix

\end{document}